\documentclass[10pt,twocolumn,letterpaper]{article}

\usepackage{cvpr} 
\usepackage{tabularx}
\usepackage{pgfplots}
\usepackage{tikz}
\usepackage{xcolor}
\usepackage{graphicx}
\usepackage{booktabs}
\usepackage{amsmath}
\usepackage{amssymb}
\usepackage{mathtools}
\usepackage{multirow}
\usepackage{array}
\usepackage{enumitem}

\definecolor{iccvblue}{rgb}{0.21,0.49,0.74}

\newcommand{\pid}{\texttt{PID}}

\usepackage[pagebackref,breaklinks,colorlinks,allcolors=iccvblue]{hyperref}

\definecolor{surfacebg}{HTML}{808E9A}

\usepackage[most]{tcolorbox}
\usepackage{xcolor}
\usepackage{tabularray}
\definecolor{surfacebg}{HTML}{808E9A}
\usepackage{makecell}
\newtcbox{\surfacebox}{
  on line,
  colback=surfacebg,
  colframe=surfacebg,
  boxrule=0pt,
  arc=1mm,
  left=4pt,right=2pt,top=1pt,bottom=1pt,
  fontupper=\bfseries\color{white}\footnotesize,
}
\def\confName{CVPR}
\def\confYear{2026}

\title{Where Should Knowledge Enter? A Layered Framework for Knowledge Infusion in Multimodal Iterative Generative Models}

\author{
Renjith Prasad$^{1}$\hspace{0.5em}
Chathurangi Shyalika$^{1}$\hspace{0.5em}
Anushka Pawar$^{2}$\\
Aahan Rathod$^{1}$\hspace{0.5em}
Amit Sheth$^{1,2}$\\[0.5em]
$^{1}$University of South Carolina\\
$^{2}$Indian AI Research Organization
}
\begin{document}
\maketitle

\begin{abstract}
Multimodal generative models produce fluent outputs but remain
unreliable when generation must respect structured, domain-specific,
or safety-critical knowledge. Existing methods incorporate knowledge
through mechanisms such as prompt augmentation, guidance, latent
editing, or fine-tuning, yet they are typically categorized by technique rather than by the component of the generative process they modify. We argue that knowledge infusion
in iterative generative models is fundamentally an
\textbf{intervention-layer problem}. Since the generative process unfolds as a
trajectory of internal states, knowledge can
act on four structurally distinct components of this process: the
input/output boundary, the transition function, the intermediate
state, and the model parameters. This maps to four intervention layers:
\textbf{surface}, \textbf{trajectory}, \textbf{latent}, and
\textbf{parametric} infusion. We instantiate the framework in
diffusion models, map representative methods to all four layers,
and derive design principles for multi-layer composition. In a controlled safety-alignment experiment using a multimodal
knowledge graph with two diffusion backbones, we implement three of
the four layers cumulatively, surface (input-side and output-side) and trajectory--latent
(mid-generation). We show empirically that each additional layer addresses failure classes that prior layers cannot reach, reducing knowledge-violating outputs by 70.97\% compared to
vanilla generation and empirically confirming the framework's complementarity prediction.

\end{abstract}
\section{Introduction}
Multimodal generative models have achieved impressive fluency in text-to-image synthesis~\cite{rombach2022highresolutionimagesynthesislatent,saharia2022photorealistictexttoimagediffusionmodels}, vision-language generation~\cite{openai2024gpt4technicalreport}, and cross-modal reasoning~\cite{lu2022learnexplainmultimodalreasoning}. Yet, fluency is not fidelity; these models remain unreliable when generation must respect domain-specific knowledge, such as anatomical constraints, scene structure, safety ontologies, or physical laws~\cite{Huang_2025,Ji_2023}. As multimodal systems move into knowledge-intensive domains~\cite{Pan_2024,Hogan_2021}, the central challenge is no longer whether they can generate, but whether they can do so \emph{consistently with what is known}.
\begin{figure*}
    \centering
    \includegraphics[width=0.8\linewidth]{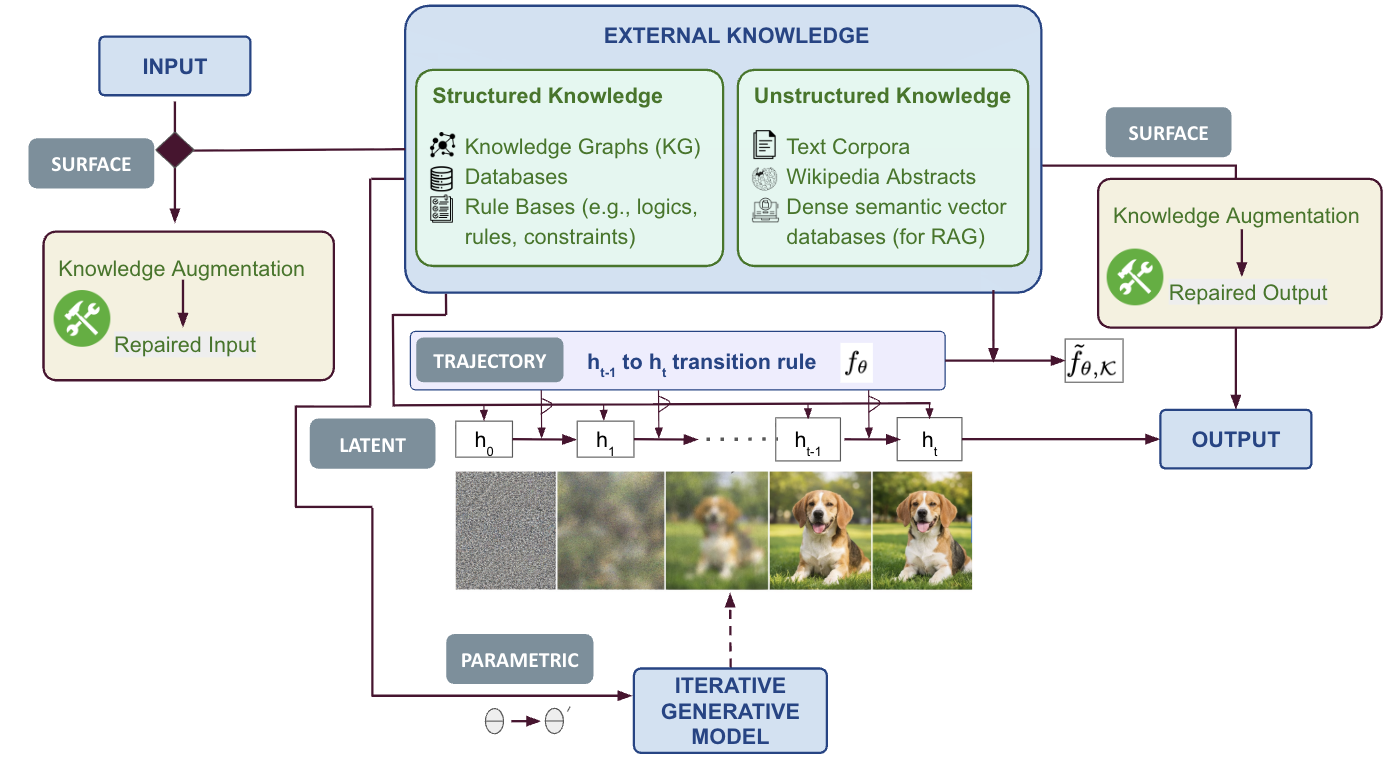}
    \caption{Four intervention layers for knowledge infusion in iterative generative models. External knowledge acts on four structurally distinct components of the generation trajectory: \textbf{surface} (input/output boundary), \textbf{trajectory} (transition rule $f_\theta$), \textbf{latent} (intermediate states $h_t$), and \textbf{parametric} (model weights $\theta$). This enables complementary coverage of prompt-level, structural, and distributional violations.}

 \label{fig:overall}
\end{figure*}

A growing body of work addresses this challenge by incorporating
external knowledge into the generation process. In practice, however, these approaches are discussed in terms of \emph{mechanism}, such as prompting and retrieval~\cite{lewis2021retrievalaugmentedgenerationknowledgeintensivenlp,gao2024retrievalaugmentedgenerationlargelanguage}, guidance~\cite{ho2022classifierfreediffusionguidance,dhariwal2021diffusionmodelsbeatgans}, latent editing~\cite{hertz2022prompt,meng2022sdeditguidedimagesynthesis}, or fine-tuning~\cite{ruiz2023dreambooth,hu2022lora,zhang2023adding}, rather than at the level of the \emph{generative process} each mechanism modifies. This makes it difficult to answer basic design questions: \emph{if a diffusion model generates an image that violates a known spatial relation, should one revise the prompt, steer the sampler, edit a latent, or retrain the model?} These interventions act at different points in the generative process and offer fundamentally different guarantees, yet no existing framework provides a principled basis for choosing among them.

We argue that knowledge infusion in iterative generative models is
fundamentally an \textbf{intervention-layer problem}. Iterative
generators like diffusion models, autoregressive decoders, and flow
models produce outputs through a trajectory of internal states:
\begin{equation}
  h_0 \;\to\; h_1 \;\to\; \cdots \;\to\; h_T = x,
  \label{eq:trajectory}
\end{equation}
where $h_0$ is an initial state, each transition
$h_t \to h_{t+1}$ is governed by a learned function
$f_\theta(\cdot, t)$, and $x$ is the generated output. This process admits four formal intervention points at which an external
knowledge signal can act, the \emph{boundary} (what enters or
exits), the \emph{transition} (the rule that advances the process),
the \emph{state} (the representation being propagated), and the
\emph{parameters} (the model that defines $f_\theta$), thereby yielding four intervention layers (Figure \ref{fig:overall}):

\begin{itemize}[leftmargin=1.5em,itemsep=2pt,topsep=4pt]
\item \textbf{Surface infusion} transforms the conditioning input
  or post-processes the output without altering the generator's
  internal dynamics.
\item \textbf{Trajectory infusion} modifies the transition function
  at inference time, steering \emph{how} the process evolves.
\item \textbf{Latent infusion} directly reshapes the intermediate
  state $h_t$, changing \emph{what} is being evolved.
\item \textbf{Parametric infusion} internalizes knowledge into the
  parameters $\theta$ or architecture, altering the
  generator itself.
\end{itemize}


\noindent These four layers are grounded in the formal structure of iterative generation: boundary, transition, state, and parameters. All methods examined in this work map to either a single layer or a composition of layers, and moving between layers changes what can be controlled, what persists across generations, and which knowledge violations can be corrected. Our framework complements the knowledge-infused learning (KiL) continuum of Sheth~et~al.~\cite{sheth2020shades}, which organizes knowledge integration in discriminative models along a shallow-to-deep axis. Complementing this view, our decomposition is defined not by depth, but by \emph{which formal component of a dynamical process} is modified. Since feedforward models lack a generation trajectory, they do not distinguish between modifying a transition rule and a propagated state; these distinctions arise only in iterative generators. The four-layer framework thus provides a complementary decomposition enabled by the trajectory structure of generation.

We instantiate the framework in diffusion models~\cite{ho2020denoising,rombach2022highresolutionimagesynthesislatent}, map representative methods to each layer, and analyze trade-offs across five axes: controllability, interpretability, persistence, computational cost, and failure-correction scope. To validate the framework, we conduct a controlled safety-alignment experiment using a multimodal knowledge graph (MMKG) as a structured knowledge source with two frozen diffusion backbones. We implement surface (input), trajectory–latent (mid-generation), and surface (output) interventions, leaving parametric infusion for future work as it requires retraining. Each added layer addresses failure classes that prior layers cannot reach, and the full multi-layer stack reduces knowledge-violating outputs by 70.97\% compared to vanilla generation, confirming that no single layer is sufficient and that principled composition yields complementary coverage.

Our key contributions are:
\begin{itemize}[leftmargin=1.5em,itemsep=2pt,topsep=3pt]
\item We formulate knowledge infusion in iterative generative models as an intervention-layer problem over the generation trajectory.

\item We introduce a four-layer framework—surface, trajectory, latent, and parametric—grounded in the formal components of iterative generation, and show that existing methods map naturally onto this space.

\item We provide a comparative analysis of the four layers along five operational axes, with representative methods mapped to each layer in diffusion models.

\item We derive three design principles for composing multi-layer knowledge infusion: matching layers to failure classes, composing for complementary coverage, and managing inter-layer interference.

\item We validate the framework empirically through a controlled safety-alignment experiment with two frozen diffusion backbones, demonstrating that each additional layer yields monotonically stronger knowledge consistency while maintaining generation quality.
\end{itemize}

\section{Problem Formulation}
\label{sec:formulation}

We formalize knowledge infusion in iterative generative models by
characterizing the generator, the knowledge source, and the
intervention-layer structure that governs how knowledge interacts
with the generation process.

\subsection{Iterative Generative Model}

We consider a multimodal generative model that produces an output $x$
conditioned on an input prompt $p$ by iterating through a sequence of
internal states:
\begin{equation}
  h_0 \;\to\; h_1 \;\to\; \cdots \;\to\; h_T = x,
  \label{eq:traj}
\end{equation}
where $h_0 = \mathrm{Init}(p, z)$ is an initial state derived from
the prompt $p$ and a stochastic seed $z$ (e.g., sampled noise in
diffusion, a start token in autoregressive decoding), and each
transition is governed by
\begin{equation}
  h_{t+1} = f_\theta(h_t,\; c_t),
  \label{eq:transition}
\end{equation}
with learnable parameters $\theta$ and a conditioning context $c_t$
that may include the step index, the prompt encoding, or other
auxiliary signals. The final state $h_T$ is either the output itself
or is decoded into $x$ by a fixed readout. This formulation
encompasses diffusion models (where $h_t$ is a noisy latent and
$f_\theta$ is the denoising step), autoregressive decoders (where
$h_t$ is a partial sequence and $f_\theta$ appends the next token),
flow-based models (where $h_t$ is a point on a learned flow
trajectory), among others.

\subsection{Knowledge Source}

Let $\mathcal{K}$ denote an external knowledge source encoding domain-specific information relevant to an underlying use case.
$\mathcal{K}$ may be structured or unstructured and take diverse forms, including knowledge graphs, ontologies, rule systems, or multimodal knowledge bases linking visual and textual concepts. We
assume that $\mathcal{K}$ induces a \emph{consistency predicate}
$\mathcal{C}(x, \mathcal{K}) \in \{0,1\}$ that evaluates whether a
generated output $x$ satisfies the constraints encoded in
$\mathcal{K}$. The goal of knowledge infusion is to increase the
probability that $\mathcal{C}(x, \mathcal{K}) = 1$ without
sacrificing output quality.

\subsection{Knowledge Infusion as Intervention}

Given the trajectory in Eq.~\ref{eq:traj}, knowledge-infusion transforms the generation process to improve consistency
with $\mathcal{K}$. The key observation is that, such a strategy must
target one or more of the formal components that define the
trajectory. An iterative generator exposes four such components:
\begin{enumerate}[leftmargin=1.8em,itemsep=3pt,topsep=4pt]
\item \textbf{The boundary:} the input $p$ that initializes
  generation and the output $x$ that results from it.
\item \textbf{The transition function:} the map
  $f_\theta(\cdot, c_t)$ that advances the process at each step.
\item \textbf{The intermediate state:} the hidden representation
  $h_t$ being propagated through the trajectory.
\item \textbf{The parameters:} the weights $\theta$ and
  architectural components that define $f_\theta$.
\end{enumerate}

\noindent Each component yields a distinct intervention layer, defined as follows. The four layers are summarized in Table~\ref{tab:layers}.

\vspace{4pt}
\noindent\textbf{Definition 1 (Surface Infusion).}
A knowledge-infusion strategy is \emph{surface} if it acts only on
the boundary of the trajectory: transforming the input
$p \mapsto p' = g_{\mathcal{K}}(p)$ before generation begins, or
post-processing the output $x \mapsto x' = r_{\mathcal{K}}(x)$ after
generation completes, without modifying $f_\theta$, any $h_t$, or
$\theta$.

\vspace{4pt}
\noindent\textbf{Definition 2 (Trajectory Infusion).}
A knowledge-infusion strategy is \emph{trajectory} if it modifies the
transition function at inference time,
$f_\theta \mapsto \tilde{f}_{\theta,\mathcal{K}}$, thereby altering
\emph{how} the state evolves from step to step, while leaving the
current state $h_t$ and the stored parameters $\theta$ unchanged.

\vspace{4pt}
\noindent\textbf{Definition 3 (Latent Infusion).}
A knowledge-infusion strategy is \emph{latent} if it directly
modifies the intermediate state,
$h_t \mapsto h_t' = \ell_{\mathcal{K}}(h_t)$, at one or more steps
during generation, changing \emph{what} is being evolved, while
leaving the transition function $f_\theta$ and the parameters
$\theta$ unchanged.

\vspace{4pt}
\noindent\textbf{Definition 4 (Parametric Infusion).}
A knowledge-infusion strategy is \emph{parametric} if it modifies the
parameters $\theta \mapsto \theta' = \theta +
\Delta\theta_{\mathcal{K}}$ or the architecture of the generator
prior to inference, altering the learned dynamics of
$f_\theta$.

\begin{table}[t]
\centering
\small
\setlength{\tabcolsep}{2.5pt}
\renewcommand{\arraystretch}{1.5}

\begin{tabular}{@{}llll@{}}
\toprule
\textbf{Layer} & \textbf{Target} & \textbf{Timing} &
  \textbf{Formal operation} \\
\midrule
\surfacebox{SURFACE} & Boundary & Pre/post & $p \mapsto p'$ or $x \mapsto x'$ \\
\surfacebox{TRAJECTORY} & Transition & Per-step & $f_\theta \mapsto \tilde{f}_{\theta,\mathcal{K}}$ \\
\surfacebox{LATENT} & State & Per-step & $h_t \mapsto h_t'$ \\
\surfacebox{PARAMETRIC} & Parameters & Pre-inference & $\theta \mapsto \theta'$ \\
\bottomrule
\end{tabular}

\renewcommand{\arraystretch}{1.0}

\caption{The four intervention layers. Each layer modifies a
  different formal component of the iterative generator.}
\label{tab:layers}
\end{table}

\section{The Four Layers: Methods and Trade-offs}
\label{sec:loci}

We examine each intervention layer in terms of its concrete
instantiation in existing methods and its profile along five
operational axes.
 \emph{(i) Controllability}: how precisely a practitioner can direct the
intervention toward a specific knowledge constraint.
 \emph{(ii) Interpretability}: whether the intervention and its effect on
the output can be directly inspected.
 \emph{(iii) Persistence}: whether the effect holds within a single
generation (transient) or across all future generations (permanent).
\emph{(iv) Computational cost}: the additional compute required, at
inference time or training time.
\emph{(v) Failure-correction scope}: which classes of knowledge
violation, prompt-level, structural, or distributional, the layer
can address, defined relative to the consistency predicate
$\mathcal{C}(x, \mathcal{K})$ from Section~\ref{sec:formulation}.
Table~\ref{tab:unified} provides a consolidated summary; ratings
are analytical assessments derived from formal layer properties, not
empirical measurements. For concreteness, we note that diffusion models instantiate
Eq.~\ref{eq:traj} as
$h_{t-1} = \frac{1}{\sqrt{\alpha_t}}
\bigl(h_t - \frac{1-\alpha_t}{\sqrt{1-\bar\alpha_t}}\,
\epsilon_\theta(h_t,t,c)\bigr) + \sigma_t z$,
where the four layers correspond to modifying $c$ (surface),
replacing $\epsilon_\theta(\cdot)$ (trajectory), editing $h_t$
(latent), and updating $\theta$ (parametric). The methods below are
drawn primarily from diffusion but the definitions apply to any
iterative generator.

\subsection{\surfacebox{SURFACE} Infusion}
\label{sec:surface}

Surface infusion operates at the boundary of the generative
trajectory. It transforms the input
$p \mapsto p' = g_{\mathcal{K}}(p)$ before generation begins, or
post-processes the output $x \mapsto x' = r_{\mathcal{K}}(x)$ after
generation completes. The internal dynamics, $f_\theta$, every
$h_t$, and $\theta$, remain untouched.

On the input side, retrieval-augmented generation
(RAG)~\cite{lewis2021retrievalaugmentedgenerationknowledgeintensivenlp}
prepends knowledge-relevant context to the prompt;
knowledge-graph-grounded prompt
rewriting~\cite{zhang2024knowgpt}
restructures conditioning using entities and relations from
$\mathcal{K}$. In autoregressive models, few-shot exemplar
selection from a knowledge base serves the same role. On the output
side, post-hoc verification checks the generated output against
$\mathcal{K}$ and either rejects or locally repairs
inconsistencies~\cite{Ji_2023}, and safety
filters~\cite{rando2022red} screen outputs against constraint
ontologies.

Surface infusion is the cheapest and most interpretable layer. Every
modification is visible in the input or output, no model access is
needed, and it composes trivially with any generator. Its corrective
reach, however, is limited to \emph{prompt-level}
violations which are errors caused by missing or ambiguous conditioning. Once
generation begins, internal dynamics may override the conditioning
signal, producing violations that surface methods cannot detect or
correct until after the output materializes.

\subsection{\surfacebox{TRAJECTORY} Infusion}
\label{sec:trajectory}

Trajectory infusion modifies the transition function at inference
time, $f_\theta \mapsto \tilde{f}_{\theta,\mathcal{K}}$, while
leaving the state $h_t$ and stored parameters $\theta$ unchanged.
The model sees the same latent but computes a different update
direction.

Classifier-free guidance~\cite{ho2022classifierfreediffusionguidance} replaces the standard noise prediction with a weighted combination of conditional and unconditional estimates, modifying the update rule to amplify prompt-aligned directions. Classifier-based
guidance~\cite{dhariwal2021diffusionmodelsbeatgans} adds an external
classifier's gradient to the score function. Diffusion posterior
sampling (DPS)~\cite{chung2022diffusion} augments the transition with
a likelihood-gradient term to enforce measurement consistency. In
autoregressive models, constrained
decoding~\cite{hokamp2017lexically} and knowledge-grounded logit
adjustment modify the next-token distribution at each step using
knowledge-derived constraints, altering the sampling rule rather than
any hidden state.

Trajectory infusion provides high controllability, as the practitioner specifies how the update rule changes, and strong persistence within a generation, since the modified transition applies at every step and enables continuous steering. Its corrective reach extends to \emph{structural} violations, including incorrect spatial relations, compositional errors, and emerging constraint violations. The cost is moderate (additional forward passes per step),
and interpretability is limited since the intervention operates in the
model's internal prediction space.

\subsection{\surfacebox{LATENT} Infusion}
\label{sec:latent}

Latent infusion directly modifies the intermediate state,
$h_t \mapsto h_t' = \ell_{\mathcal{K}}(h_t)$, at one or more steps
during generation. The transition function $f_\theta$ and the
parameters $\theta$ are unchanged, the model applies its usual
dynamics, but to an altered state.

Prompt-to-Prompt editing~\cite{hertz2022prompt} overwrites
cross-attention maps within the denoising network, surgically
redirecting which spatial regions attend to which semantic concepts.
SDEdit~\cite{meng2022sdeditguidedimagesynthesis} re-noises a
reference image to produce a modified $h_0'$ and runs standard
denoising from there---a latent intervention at $t\!=\!0$. Latent
projection~\cite{gandikota2023erasing} maps intermediate
representations onto or away from knowledge-defined subspaces. In
autoregressive models, activation
editing~\cite{li2023inference} and representation
engineering~\cite{zou2022representation} directly modify internal
hidden states during the forward pass.

Latent infusion offers the most fine-grained controllability. The
practitioner can target specific features, spatial regions, or
semantic dimensions of the hidden state. Its corrective reach also covers
\emph{structural} violations, but through a different mechanism than trajectory infusion: rather than changing the direction of travel, it moves the traveler to a different position. The critical
limitation is persistence: latent edits are pointwise perturbations,
and the unmodified dynamics at subsequent steps may attenuate or
override them. This challenge has motivated step-range-controlled
attention injection~\cite{hertz2022prompt}, per-step null-text
optimization~\cite{mokady2023null}, and noise-level-dependent
faithfulness--realism
tradeoffs~\cite{meng2022sdeditguidedimagesynthesis}.

\subsection{\surfacebox{PARAMETRIC} Infusion}
\label{sec:core}

Parametric infusion modifies the parameters
$\theta \mapsto \theta' = \theta + \Delta\theta_{\mathcal{K}}$ or
the architecture of the generator prior to inference. Unlike the
three inference-time layers, parametric infusion permanently changes the
generator's learned dynamics.

DreamBooth~\cite{ruiz2023dreambooth} fine-tunes the full denoising
network on concept-specific images, embedding new knowledge into
$\theta$. LoRA~\cite{hu2022lora} injects trainable low-rank matrices
into attention layers with minimal overhead.
ControlNet~\cite{zhang2023adding} adds a parallel encoder branch that injects spatial conditioning, permanently extending the model's input interface. Concept
erasure~\cite{gandikota2023erasing} fine-tunes $\theta$ to remove
specific concepts from the generative distribution entirely. In
autoregressive models, knowledge-grounded fine-tuning and adapter
injection~\cite{hu2022lora} serve the same role.

Parametric infusion is the only layer with permanent persistence, with knowledge encoded in $\theta$ and applied to all future generations without per-instance cost. Its corrective reach extends to
\emph{distributional} violations, systematic biases or missing
concepts in the model's learned distribution that no inference-time
intervention can address. The cost is highest (training compute, data curation), and flexibility is lowest, as updating or retracting knowledge requires retraining.

\subsection{Borderline Cases}
\label{sec:borderline}

Not every method maps to a single layer.
Attend-and-Excite~\cite{chefer2023attend} computes a loss
over attention maps (a function of $h_t$) and backpropagates through
it to update $h_t$: the loss design is a trajectory-layer choice
(it shapes the update rule), while the gradient step that modifies
$h_t$ is a latent-layer operation. We classify it as a
trajectory--latent composition. Similarly,
DPS~\cite{chung2022diffusion} augments the score function with a
likelihood gradient, primarily a trajectory intervention, but the
gradient is applied directly to $h_t$, producing a secondary latent
effect. The framework accommodates these cases as multi-layer
compositions with a primary and secondary layer, rather than
demanding single-label classification.
\subsection{Cross-Layer Insights}
\label{sec:insights}

\paragraph{Persistence separates trajectory from latent.}
Both layers intervene during generation with comparable
controllability, but trajectory infusion modifies the map
$f_\theta \mapsto \tilde{f}_{\theta,\mathcal{K}}$ continuously
across its active window, while latent infusion perturbs the
argument $h_t$ at discrete points. The unmodified dynamics may
attenuate latent edits; they cannot attenuate a trajectory
modification that is reapplied at every step. This persistence gap
is the principal operational reason to distinguish the two layers.

\paragraph{No single layer dominates.}
Surface infusion is cheap and interpretable but shallow in corrective
reach. Parametric infusion is permanent and broad but expensive and
inflexible. Trajectory and latent infusion offer fine-grained
inference-time control but are transient. This complementarity
follows from the fact that each layer targets a different formal
component of the generator.

\begin{table*}[t]
\centering
\small

\vspace{4pt}
\setlength{\tabcolsep}{2.5pt}
\begin{tabular}{@{}lcccccp{5.2cm}@{}}
\toprule
\textbf{Layer} & \textbf{Controllability} &
  \textbf{Interpretability} & \textbf{Persistence} & \textbf{Cost}
  & \textbf{Failure Scope} & \textbf{Representative Methods} \\
\midrule
\surfacebox{SURFACE}
  & Low & High & Transient & Low & Prompt-level
  & RAG~\cite{lewis2021retrievalaugmentedgenerationknowledgeintensivenlp},
    KG-grounded
    prompting~\cite{zhang2024knowgpt},
    output filtering~\cite{rando2022red} \\[4pt]
\surfacebox{TRAJECTORY}
  & High & Low & \makecell{Transient \\ (continuous)} & Moderate &
  Structural
  & Classifier- and classifier-free guidance~\cite{ho2022classifierfreediffusionguidance,dhariwal2021diffusionmodelsbeatgans},
    DPS~\cite{chung2022diffusion},
    constrained decoding~\cite{hokamp2017lexically} \\[4pt]
\surfacebox{LATENT}
& High & Low & \makecell{Transient \\ (attenuating)\textsuperscript{$\dagger$}} & Moderate & Structural
  & Prompt-to-Prompt~\cite{hertz2022prompt},
    SDEdit~\cite{meng2022sdeditguidedimagesynthesis},
    latent projection~\cite{gandikota2023erasing},
    activation editing~\cite{li2023inference} \\[4pt]
\surfacebox{PARAMETRIC}
  & Low & Low & Permanent & High & Distributional
  & DreamBooth~\cite{ruiz2023dreambooth},
    LoRA~\cite{hu2022lora},
    ControlNet~\cite{zhang2023adding},
    concept erasure~\cite{gandikota2023erasing} \\
\bottomrule
\end{tabular}
\caption{Unified summary of the four intervention layers. Ratings
  are analytical assessments derived from the formal properties of
  each layer (Section~\ref{sec:formulation}), not empirical
  measurements; empirical validation is presented in
  Section~\ref{sec:empirical}. Failure classes are defined in
  Section~\ref{sec:insights}.}
\label{tab:unified}
\vspace{2pt}
{\footnotesize \textsuperscript{$\dagger$}Latent edits are subject
  to attenuation by subsequent unmodified
  dynamics~\cite{hertz2022prompt,mokady2023null}, making their
  effective persistence weaker than trajectory infusion at the same
  number of active steps.}
\end{table*}

\begin{figure*}[t]
    \centering
    \includegraphics[width=\textwidth]{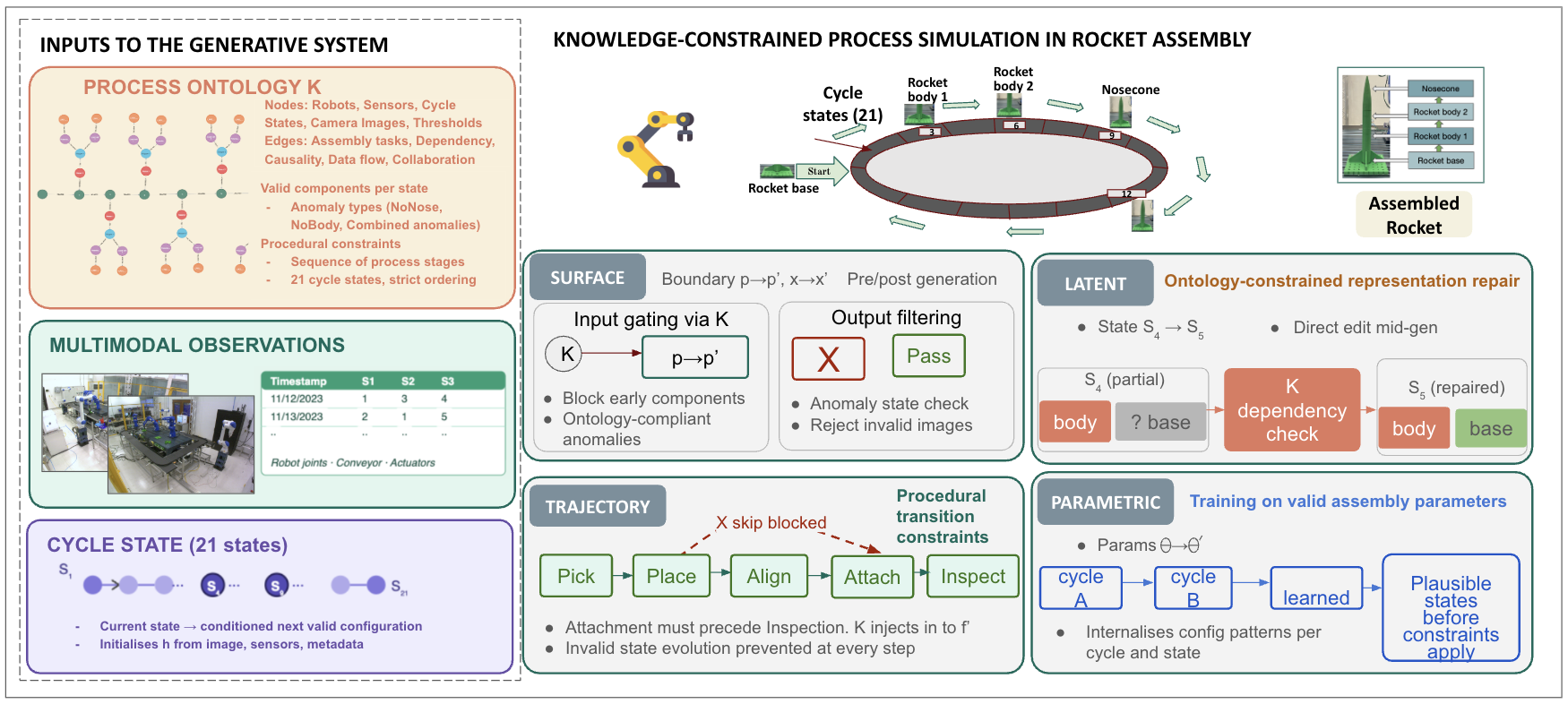}
    \caption{
    Ontology-guided knowledge infusion for process simulation in a rocket
    assembly pipeline. The process ontology serves as structured knowledge
    $\mathcal{K}$ and constrains generation at four layers:
    \emph{surface} gating at input/output boundaries, \emph{trajectory}
    constraints over valid procedural transitions, \emph{latent} repair of
    intermediate structural inconsistencies, and \emph{Parametric} learning from
    valid assembly trajectories. This illustrates the framework in a
    highly structured domain where knowledge is tied to temporal ordering
    and process physics.
    }
    \label{fig:rocket}
\end{figure*}

\section{Use Cases}
\label{sec:usecases}

We instantiate the framework in two domains with differing knowledge types, architectures, and failure modes. This demonstrates that our framework generalizes across open-ended generation and structured process simulation.
\subsection{Knowledge-Constrained Process Simulation in Rocket Assembly}

We ground the proposed framework in a smart manufacturing testbed based on a rocket assembly pipeline from the Future Factories (FF) lab \cite{harik2024analog}. The process consists of assembly cycles, each divided into 21 discrete cycle states ($S$) representing sequential operations such as component placement, alignment, and attachment. Each cycle produces a multimodal record comprising synchronized images from multiple cameras and sensor measurements (e.g., robot joint angles, conveyor states, and actuator signals) \cite{prasad2024assemai, shyalika2025nsf}.

A key characteristic of this process is that valid system evolution is highly structured, which means only specific components can be present at each cycle state. Transitions between states must follow predefined procedural constraints. These constraints, along with anomaly definitions (e.g., missing nose cone or body segments), are encoded in a process ontology that serves as the knowledge source $K$.

We consider a generative simulation setting in which the model predicts future process states (e.g., next visual frame or structured configuration) given the current state. Formally, starting from an initial state $h_0$ derived from the current cycle (image, sensors, and metadata), the model generates a trajectory:
\[
h_0 \rightarrow h_1 \rightarrow \cdots \rightarrow h_T,
\]
where each state corresponds to a candidate configuration of the assembly process.

\paragraph{Inputs and Outputs.}
The input consists of (i) the current cycle state (e.g., $S_4$ or $S_9$), (ii) synchronized multimodal observations (images and sensor values), and (iii) ontology constraints specifying valid components and transitions. The output is a sequence of images representing the predicted future states of the assembly process.

\paragraph{How Knowledge Acts in This System (Figure \ref{fig:rocket})}
\leavevmode\\
\textbf{Surface infusion.} At the boundary, ontology knowledge is used to constrain valid initializations and outputs. For example, given a current cycle state, the ontology restricts which components can appear (e.g., a nose component cannot be present before a certain stage). Generated outputs that violate these rules (e.g., predicting a component too early in the sequence) are filtered and rejected.

\textbf{Trajectory infusion.} The assembly process follows strict procedural transitions (e.g., component attachment must precede inspection). These transition rules are injected into the generation process to prevent invalid state evolution, such as skipping required steps or generating incompatible component configurations. This ensures that predicted trajectories respect the temporal ordering of the manufacturing process.

\textbf{Latent infusion.} Intermediate states may contain localized inconsistencies, such as partial and missing components in the generated representation. These are corrected by enforcing ontology constraints at the representation level, for instance ensuring that if a body segment is present, its corresponding supporting components must also exist. This enables fine-grained correction of structural errors during generation.

\textbf{Parametric infusion.} The generative model is trained on trajectories derived from valid assembly cycles, allowing it to internalize common patterns such as typical component configurations at each cycle state and the progression of assembly steps. This enables the model to produce plausible states even before explicit constraints are applied.

\paragraph{Discussion.}
This use case highlights that knowledge in manufacturing is tightly coupled to process structure: constraints arise from physical assembly rules and temporal ordering rather than abstract semantics. Surface constraints eliminate invalid configurations at the boundaries, trajectory constraints enforce correct sequencing, latent corrections fix local structural violations, and parametric learning captures recurring process dynamics. These mechanisms ensure that generated trajectories remain consistent with both the physical and procedural requirements of the assembly pipeline.

\subsection{Multi-Layer Safety Alignment}
\label{sec:safety}

\begin{figure*}[t]
    \centering
    {\color{red}\small\textbf{Content advisory:} Includes violence, sexually explicit imagery, and harmful stereotypes.\par}
    \vspace{2pt}
    \includegraphics[width=\textwidth]{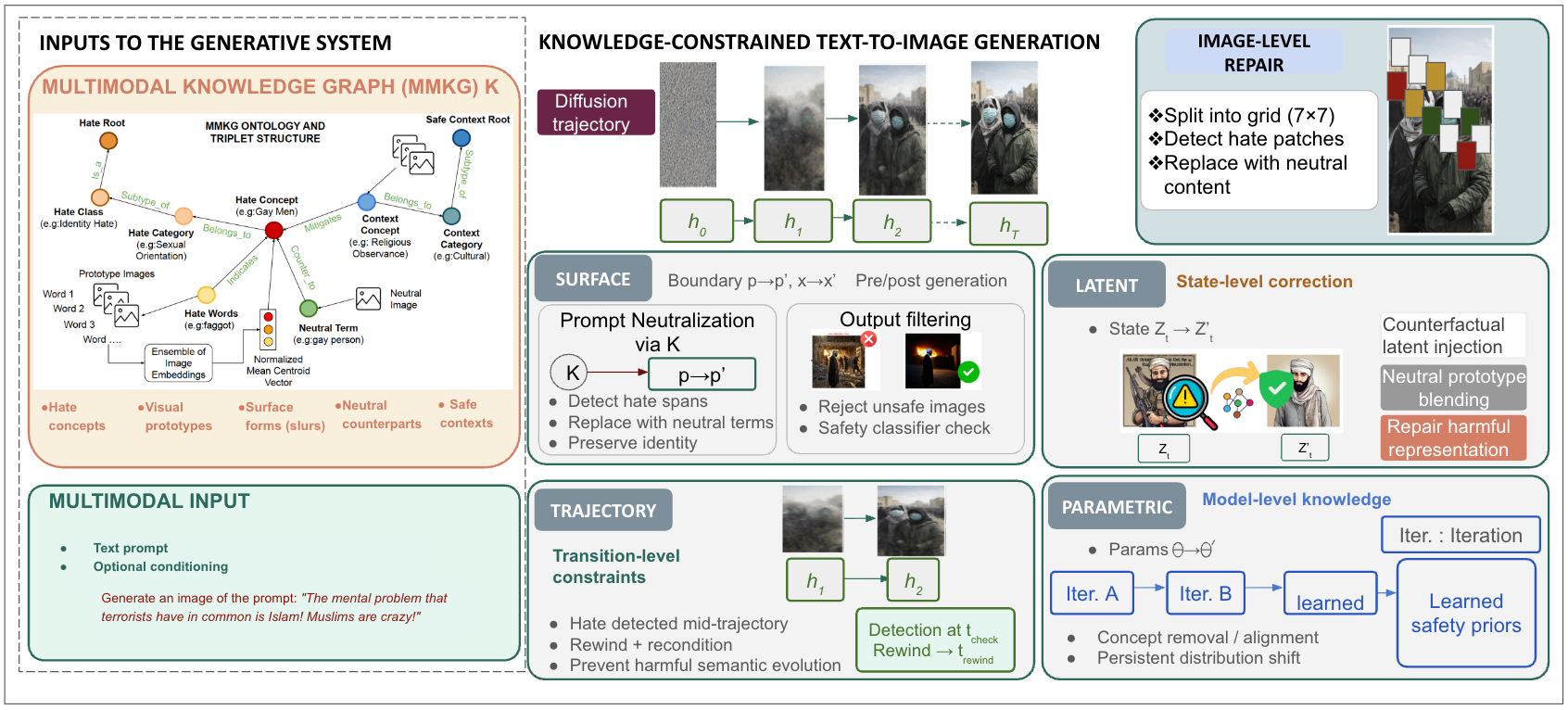}
    \caption{
    Multi-layer knowledge infusion for safety alignment in text-to-image
    diffusion using a multimodal knowledge graph (MMKG). Input-side
    \emph{surface} infusion neutralizes unsafe prompt spans using graph-linked
    counterparts; \emph{trajectory--latent} intervention detects harmful
    activation during denoising, rewinds and repairs the latent, and resumes
    sampling with neutral conditioning; output-side \emph{surface} infusion
    identifies unsafe local regions and repairs them via targeted inpainting.
    MMKG provides a shared control signal across all active layers.
    }
    \label{fig:safety}
\end{figure*}

We instantiate the framework for safety alignment in text-to-image
diffusion, where a multimodal knowledge graph (MMKG) serves as the
structured knowledge source $\mathcal{K}$
(Figure~\ref{fig:safety}).

\paragraph{Knowledge source.}
The MMKG encodes approximately $10^4$ textual nodes and $10^3$
visual prototypes as a typed graph. Hate concepts are linked to
surface forms (slurs, trope phrases), CLIP ViT-L/14 visual
centroids, safe-context concepts, and explicit neutral counterparts
via typed relations (\texttt{Indicates}, \texttt{Counter\_to},
\texttt{Mitigates}). Consistency predicate
$\mathcal{C}(x,\mathcal{K})=1$ holds if and only if the generated image $x$
contains no visual realization of any hate concept in $\mathcal{K}$,
as determined by an external multimodal safety classifier.

\paragraph{Layer instantiation.}
Using frozen SDXL and SD-v1.5 backbones, we activate three layers
cumulatively:

\noindent\textbf{Surface (input-side):} MMKG-guided prompt
neutralization detects hate-concept surface forms via
obfuscation-tolerant graph lookup and a RoBERTa-based clause-level
stance classifier, replacing hateful spans with MMKG-provided
neutral terms while preserving identity mentions:
$p \mapsto p' = g_{\mathcal{K}}(p)$.

\noindent\textbf{Trajectory--latent (mid-generation):} At a late
denoising step ($t_{\text{check}} \approx 0.9T$), a decoded preview
is scored against MMKG hate prototypes via CLIP. If activation
exceeds a calibrated threshold, the system rewinds the latent to
$t_{\text{rewind}} \approx 0.3T$ by re-noising and blending with a
neutral reference latent (latent: $h_t \mapsto h_t'$), then resumes
with MMKG-derived neutral conditioning (trajectory:
$f_\theta \mapsto \tilde{f}_{\theta,\mathcal{K}}$).

\noindent\textbf{Surface (output-side):} A $7\times7$ CLIP patch
grid scores local image regions against MMKG hate and safe-context
prototypes. Patches closer to hate than safe prototypes are locally
inpainted using MMKG-derived neutral prompts:
$x \mapsto x' = r_{\mathcal{K}}(x)$.

\noindent All modules are training-free with respect to the
diffusion backbone. Parametric infusion is not evaluated in this
experiment as it requires retraining; the design holds backbones
frozen to isolate inference-time layer composition.

\paragraph{Baselines and metrics.}
We evaluate on the Detonate benchmark~\cite{prasad2025detonate}
(25K prompts) against vanilla generation,
SAFREE~\cite{yoon2024safree} (embedding-space steering), and
SLD~\cite{schramowski2023safe} (safety-guided diffusion). We report
toxicity (fraction flagged as hateful; lower is better), CLIP score
(text--image alignment; higher is better), and AQI (aesthetic
quality; higher is better), generating 4 samples per prompt and
reporting means.

\section{Empirical Results}
\label{sec:empirical}

We evaluate the multi-layer safety system described in
Section~\ref{sec:safety}. Table~\ref{tab:empirical} reports results
for both backbones; the pattern is consistent across architectures.

\begin{table}[t]
\centering
\small
\setlength{\tabcolsep}{4pt}

\begin{tabular}{@{}lcccc@{}}
\toprule
\textbf{Layer} &
\textbf{Control} &
\textbf{Interp.} &
\textbf{Persistence} &
\textbf{Cost} \\
\midrule
Surface     & Low  & High & Low  & Low \\
Trajectory  & High & Low  & High & Med \\
Latent      & High & Low  & Med  & Med \\
Parametric  & Low  & Low  & High & High \\
\bottomrule
\end{tabular}
\end{table}
\begin{table}[t]
\centering
\small

\vspace{4pt}
\setlength{\tabcolsep}{3.5pt}
\begin{tabular}{@{}lcccccc@{}}
\toprule
& \multicolumn{3}{c}{\textbf{SDXL}} &
  \multicolumn{3}{c}{\textbf{SD-v1.5}} \\
\cmidrule(lr){2-4} \cmidrule(lr){5-7}
\textbf{Config.} &
  \textbf{Tox$\downarrow$} & \textbf{CLIP$\uparrow$} &
  \textbf{AQI$\uparrow$} &
  \textbf{Tox$\downarrow$} & \textbf{CLIP$\uparrow$} &
  \textbf{AQI$\uparrow$} \\
\midrule
Vanilla
  & .31 & .310 & .23 & .28 & .330 & .22 \\
SAFREE~\cite{yoon2024safree}
  & .22 & .305 & .28 & .21 & .320 & .27 \\
SLD~\cite{schramowski2023safe}
  & .18 & .320 & .31 & .17 & .335 & .30 \\
\midrule
Surface (input)
  & .17 & .330 & .32 & .16 & .345 & .31 \\
+Traj.--latent
  & .11 & .340 & .37 & .10 & .360 & .36 \\
+Surface (output)
  & \textbf{.09} & .335 & .36
  & \textbf{.08} & .350 & .35 \\
\bottomrule
\end{tabular}
\caption{Multi-layer safety alignment on Detonate. Each row adds
  one layer cumulatively. All methods use frozen backbones.}
\label{tab:empirical}
\end{table}

\paragraph{Finding 1: Different layers correct different failure classes.}
Surface input infusion reduces toxicity from 0.31 to 0.17 (SDXL) by
neutralizing prompt-level triggers, but 17\% of outputs remain
toxic, reflecting structural violations where pretrained weights
hallucinate hateful motifs despite a clean prompt. Trajectory--latent
intervention catches these mid-generation failures (0.11).
Output-side surface infusion removes residual artifacts surviving
both prior layers (0.09). This layered pattern matches the
framework's prediction that prompt-level, structural, and residual
violations require distinct layers
(Section~\ref{sec:insights}).

\paragraph{Finding 2: Multi-layer composition outperforms any
single layer.}
The full stack (0.09) halves the best single-layer toxicity (0.17)
and outperforms both baselines (SAFREE: 0.22, SLD: 0.18) while
maintaining higher CLIP and AQI, confirming complementary coverage
(Section~\ref{sec:insights}).

\paragraph{Finding 3: Shared knowledge structure prevents
inter-layer conflict.}
All layers query the same MMKG, with shared hate concepts, neutral
counterparts, and a consistency predicate driving prompt neutralization,
trajectory steering, and output repair. This shared grounding
ensures complementary rather than conflicting effects, supporting
the principle that structured knowledge should function as a
unified control signal across the trajectory.

\section{Discussion and Future work}
\label{sec:conclusion}

\paragraph{Limitations.}
The trade-off ratings in Table~\ref{tab:unified} are analytical; controlled empirical measurement of controllability, interpretability, and persistence across standardized benchmarks would strengthen these claims. Our treatment of autoregressive and flow-based generators is less developed than the diffusion setting, and extending the case study to these families would strengthen generality. Finally, our empirical validation covers a single task (safety), knowledge modality (MMKG), and model family (latent diffusion); generalization to other domains requires further study.

\paragraph{Open problems.}
The framework surfaces research questions that, to our knowledge,
have not been explicitly posed.

\emph{Adaptive layer selection.}
Given a knowledge source $\mathcal{K}$ and a detected violation of $\mathcal{C}(x, \mathcal{K})$, can the appropriate intervention layer be selected automatically? This requires learning a mapping from violation type to intervention target, a diagnostic problem that the four-layer vocabulary makes precise but does not solve.

\emph{Inter-layer interference.}
Our experiment used a shared MMKG to ensure coherent multi-layer composition (Finding~3), but independently designed layers may conflict, as trajectory guidance and latent edits push in different directions. Characterizing when
composition is synergistic versus destructive, and developing
scheduling or joint-optimization strategies, is an open problem with
practical consequences for modular knowledge-grounded systems.

\emph{Knowledge-aware trajectory design.}
Current generators are trained without awareness of the knowledge constraints they must satisfy. Designing objectives that make trajectories more amenable to inference-time infusion is a promising direction, for example by learning disentangled latent dimensions aligned with knowledge-graph relations.

\emph{Evaluation protocol.}
The framework's comparative claims call for a standardized benchmark.
A knowledge-consistency evaluation could pair a structured knowledge
source $\mathcal{K}$ with generation prompts whose outputs must
satisfy specific relational or attributive constraints derived from
$\mathcal{K}$. Violation rates under the consistency predicate
$\mathcal{C}(x, \mathcal{K})$ could then be compared across
single-layer and multi-layer configurations, controlling for output
quality via standard metrics (FID, CLIPScore). Designing such a
benchmark is a concrete next step toward comprehensive empirical
validation.


\section{Conclusion}
We formulate knowledge infusion in iterative generative models as an intervention-layer problem and introduce a four-layer framework grounded in the formal components of generation. The framework provides a principled design space for analyzing and composing knowledge-integration strategies. We instantiate it in diffusion models and validate its predictions in a controlled safety-alignment setting, showing that different layers address distinct failure classes and that multi-layer composition yields complementary gains. More broadly, the framework shifts the question from which knowledge source to use to where and how that knowledge should enter the generative process. An important direction for future work is adaptive selection and coordination of intervention layers based on violation type, model architecture, and deployment constraints.

\bibliographystyle{ieeenat_fullname}
\bibliography{main}

\end{document}